\documentclass{article}

\usepackage[final, nonatbib]{neurips_2020}

\usepackage[utf8]{inputenc} 
\usepackage[T1]{fontenc}    
\usepackage{hyperref}       
\usepackage{url}            
\usepackage{booktabs}       
\usepackage{amsfonts}       
\usepackage{nicefrac}       
\usepackage{microtype}      

\usepackage{algorithm}
\usepackage{algpseudocode}
\usepackage{xcolor}

\usepackage{graphicx}
\usepackage{caption}
\usepackage{subfigure}

\def\bzero{\boldsymbol{0}}

\usepackage{amsmath,amssymb,amsfonts}

\def\bb{\boldsymbol{b}}
\def\bc{\boldsymbol{c}}

\def\bo{\boldsymbol{o}}

\def\bu{\boldsymbol{u}}
\def\bv{\boldsymbol{v}}

\def\bx{\boldsymbol{x}}
\def\by{\boldsymbol{y}}

\def\bA{\boldsymbol{A}}
\def\bB{\boldsymbol{B}}
\def\bC{\boldsymbol{C}}
\def\bD{\boldsymbol{D}}

\def\bI{\boldsymbol{I}}

\def\bO{\boldsymbol{O}}

\def\bQ{\boldsymbol{Q}}
\def\bR{\boldsymbol{R}}

\def\bX{\boldsymbol{X}}
\def\bY{\boldsymbol{Y}}
\def\bZ{\boldsymbol{Z}}
\def\thick#1{\hbox{\rlap{$#1$}\kern0.25pt\rlap{$#1$}\kern0.25pt$#1$}}

\def\bbeta{\boldsymbol{\beta}}

\def\btheta{\boldsymbol{\theta}}

\def\bmu{\boldsymbol{\mu}}

\def\bomega{\boldsymbol{\omega}}

\def\bSigma{\boldsymbol{\Sigma}}

\def\bOmega{\boldsymbol{\Omega}}
\def\thick#1{\hbox{\rlap{$#1$}\kern0.25pt\rlap{$#1$}\kern0.25pt$#1$}}
\def\smbalpha{{\thick{\scriptstyle{\alpha}}}}

\def\smbalpha{\widehat{\smbalpha}}

\def\hbar{{\overline h}}

\def\Ssc{{\mathcal S}}

\def\smhalf{{\textstyle{\frac{1}{2}}}}

\def\simind{\stackrel{{\tiny \mbox{ind.}}}{\sim}}
\def\blockdiagdum{\mathop{\mbox{\rm blockdiag}}}
\def\blockdiag#1{\blockdiagdum_{#1}}

\def\stackdum{\mathop{\mbox{\rm stack}}}
\def\stack#1{\stackdum_{#1}}

\def\beq{\begin{equation}}
\def\eeq{\end{equation}}
\def\bev{\begin{verbatim}}
\def\eev{\end{verbatim}}

\def\lboxit#1{\vbox{\hrule\hbox{\vrule\kern6pt
      \vbox{\kern6pt#1\kern6pt}\kern6pt\vrule}\hrule}}

\def\thickboxit#1{\vbox{{\hrule height 1mm}\hbox{{\vrule width 1mm}\kern6pt
          \vbox{\kern6pt#1\kern6pt}\kern6pt{\vrule width 1mm}}
               {\hrule height 1mm}}}
\def\colthickboxit#1{\vbox{{\textcolor{brown}{\hrule height 3mm}}
          \hbox{{\textcolor{brown}{\vrule width 3mm}}\kern6pt
          \vbox{\kern6pt#1\kern6pt}\kern6pt{\textcolor{brown}{\vrule width 3mm}}}
               {\hrule height 5mm}}}

\def\beq{\begin{eqnarray}}
\def\eeq{\end{eqnarray}}
\def\beqn{\begin{eqnarray*}}
\def\eeqn{\end{eqnarray*}}

\def\sigsqeps{\sigeps^2}

\def\bse{\begin{eqnarray*}}
\def\ese{\end{eqnarray*}}
\def\raybe{\begin{eqnarray}}
\def\rayee{\end{eqnarray}}

\def\fat#1{\hbox{\rlap{$#1$}\kern0.25pt\rlap{$#1$}\kern0.25pt$#1$}}

\def\thickarrow{\longleftarrow}

\def\sigsqeps{\sigma^2_{\varepsilon}}

\def\bse{\begin{eqnarray*}}
\def\ese{\end{eqnarray*}}
\def\raybe{\begin{eqnarray}}
\def\rayee{\end{eqnarray}}

\def\pe+{p_{{\scriptscriptstyle{\rm E}+}}}
\def\pg+{p_{{\scriptscriptstyle{\rm G}+}}}

\def\exerDBtags#1#2{\null}
{\begin{list}
{\bulletcolour$\bullet$}
{\setlength{\leftmargin}{30mm}\setlength{\itemsep}{0.5ex}}\sf}
{\end{list}\normalsize}
{\begin{list}{\small\bulletcolour$\bullet$}
{\setlength{\leftmargin}{0.75in}\setlength{\itemsep}{0.5ex}}\small}
{\end{list}\normalsize}
\def\simind{\stackrel{{\tiny \mbox{ind.}}}{\sim}}
\def\btheta{\boldsymbol{\theta}}

\def\bbeta{\boldsymbol{\beta}}
\def\bSigma{\boldsymbol{\Sigma}}
\def\bCov{\textbf{Cov}}
\def\sigsqeps{\sigma_{\varepsilon}^{2}}

\newtheorem{result}{\textbf{Result}}
\def\bzero{\textbf{0}}

\def\by{\textbf{\textit{y}}}
\def\bu{\textbf{\textit{u}}}
\def\bx{\textbf{\textit{x}}}
\def\bo{\textbf{\textit{o}}}
\def\bA{\textbf{\textit{A}}}
\def\bB{\textbf{\textit{B}}}
\def\bC{\textbf{\textit{C}}}
\def\bD{\textbf{\textit{D}}}

\def\bI{\textbf{\textit{I}}}

\def\bO{\textbf{\textit{O}}}

\def\bQ{\textbf{\textit{Q}}}
\def\bR{\textbf{\textit{R}}}

\def\bX{\textbf{\textit{X}}}
\def\bY{\textbf{\textit{Y}}}
\def\bZ{\textbf{\textit{Z}}}
\def\bveco{\boldsymbol{b}_1}
\def\bvect{\boldsymbol{b}_2}
\def\bveci{\boldsymbol{b}_i}
\def\bvecm{\boldsymbol{b}_m}

\def\Bmato{\bB_1}
\def\Bmatt{\bB_2}
\def\Bmati{\bB_i}
\def\Bmatm{\bB_m}

\def\Bmatdoto{\bBdot_1}
\def\Bmatdott{\bBdot_2}
\def\Bmatdoti{\bBdot_i}
\def\Bmatdotm{\bBdot_m}

\def\AtLev{\bA}
\def\TinySolveTwoLevelSparseLeastSquares{\textsc{\scriptsize STLSLS}}
\def\SolveTwoLevelSparseLeastSquares{\textsc{\footnotesize STLSLS}}
\def\LargerSolveTwoLevelSparseLeastSquares{\textsc{\normalsize STLSLS}}
\def\CPU{\texttt{\footnotesize GPyT-CPU}}
\def\GPU{\texttt{\footnotesize GPyT-GPU}}
\def\SEB{\texttt{\footnotesize sEB}}
\def\smalldot{\mbox{\fontsize{0.1mm}{0.5em}\selectfont{$\bullet$}}}
\def\bBdot{\overset{\ \smalldot}{\bB}}
\def\Bmati{\bB_i}
\def\Bmatdoti{\bBdot_i}
\def\bveci{\boldsymbol{b}_i}
\def\xveco{\bx_1}
\def\xvectCi{\bx_{2,i}}
\def\AUoo{\bA^{11}}
\def\AUttCi{\bA^{22,i}}
\def\AUotCi{\bA^{12,i}}
\def\AUotCo{\bA^{12,1}} \def\AUotCoT{\bA^{12,1\,T}}
\def\AUotCt{\bA^{12,2}} \def\AUotCtT{\bA^{12,2\,T}}
\def\AUotCi{\bA^{12,i}} 
\def\AUotCm{\bA^{12,m}} \def\AUotCmT{\bA^{12,m\,T}}
\def\AUttCo{\bA^{22,1}}
\def\AUttCt{\bA^{22,2}}
\def\AUttCi{\bA^{22,i}}
\def\AUttCm{\bA^{22,m}}
\def\bigX{{\LARGE\mbox{$\times$}}}
\def\xvectCi{\bx_{2,i}}
\def\Cmatoi{\bC_{1i}}
\def\cveczi{\bc_{0i}}
\def\cvecoi{\bc_{1i}}
\def\cvecti{\bc_{2i}}
\def\Cmatzi{\bC_{0i}}
\def\Cmatoi{\bC_{1i}}
\def\Cmatti{\bC_{2i}}
\def\nadj{{\tilde n}}
\def\stackdum{\mathop{\mbox{\rm stack}}}
\def\stack#1{\stackdum_{#1}}
\def\blockdiagdum{\mathop{\mbox{\rm blockdiag}}}
\def\blockdiag#1{\blockdiagdum_{#1}}

\def\OmegaAtwoTwo{\bOmega_4}

\title{Fast Physical Activity Suggestions: \\ 
Efficient Hyperparameter Learning in Mobile Health}

\author{%
  Marianne Menictas \\
  Harvard University \\
  Cambridge, MA 02138 \\
  \texttt{marianne\_menictas@fas.harvard.edu} \\
   \And
   Sabina Tomkins \\
   Stanford University \\
   Stanford, CA 94305 \\
   \texttt{stomkins@stanford.edu} \\
   \AND
   Susan A. Murphy \\
   Harvard University \\
   Cambridge, MA 02138 \\
   \texttt{samurphy@fas.harvard.edu} \\
}

\begin{document}

\maketitle

\begin{abstract}
  Users can be supported to adopt healthy behaviors, such as 
regular physical activity, via relevant and timely suggestions 
on their mobile devices. Recently, reinforcement learning algorithms 
have been found to be effective for learning the optimal context under which 
to provide suggestions. However, these algorithms are not necessarily designed for the 
constraints posed by mobile health (mHealth) settings, that they be efficient, domain-informed 
and computationally affordable. 
We propose an algorithm for providing physical activity suggestions in  mHealth 
settings. Using domain-science, we formulate a contextual bandit algorithm 
which makes use of a linear mixed effects model. We then introduce a  procedure 
to efficiently perform hyper-parameter updating, using 
  far less computational resources than competing approaches. 
 Not only is our 
  approach computationally efficient, it is also easily implemented with closed 
  form matrix algebraic updates and we show improvements over state of the art 
  approaches both in speed and accuracy of up to 99\% and 56\% respectively.
\end{abstract}

\section{Introduction}
Physical activity can effectively reduce the risk of severe health problems, 
such as heart disease, yet many people
at-risk fail to exercise regularly \cite{berlin1990meta,billinger2014physical}.
Mobile health (mHealth) applications 
which provide users with regular nudges over their mobile devices can 
offer users with convenient and steady support to adopt healthy physical 
activity habits and have been shown to 
increase physical activity drastically \cite{klasnja2019efficacy}. 
However, to realize their potential, algorithms to provide physical activity suggestions 
must be efficient, domain-specific, and computationally affordable. 

We present an algorithm designed for an mHealth study in which an
online Thompson Sampling (TS) contextual bandit algorithm is used to personalize
the delivery of physical activity suggestions \cite{tomkins2020intelligentpooling}.
These suggestions are intended to increase near time physical activity.
The personalization occurs via: \emph{(i)} the user's current context, which 
is used to decide whether to deliver a suggestion; and \emph{(ii)} random effects, 
which are used to learn user and time specific parameters that encode the 
influence of the context. The user and time specific parameters are
modeled in the reward function (the mean of the reward conditional on
context and action). To learn these parameters, information is pooled across 
users and time dynamically, combining TS with a Bayesian 
random effects model. 

The contributions of this paper are: \emph{(i)} the development of a TS 
algorithm involving fast empirical Bayes (EB) fitting of a Bayesian random effects model for 
the reward; \emph{(ii)} the EB fitting procedure only requires storage and 
computing at each iteration on the order of $\mathcal{O}(m_1 m_2^3)$, 
where $m_1$ is the larger dimension of the two grouping mechanisms considered. 
For example, $m_1$ may represent the number of users and $m_2$ the time points 
or vice-versa; and \emph{(iii)} Our approach reduces the running time over  
state-of-art methods, and critically, does not require advanced hardware.

\section{Methods: Hyperparameter learning for mHealth reward functions}

\subsection{Problem Setting}\label{Sec:probSetting}

At each time, $t$, on each user, $i$, a vector of context variables,
$\bX_{i t}$, is observed. An action, $A_{i t}$, is selected.
We consider $K$ actions, where $K \in \mathbb{N}$. Subsequently a 
real-valued reward, $Y_{it}$ is observed. This continues for $t = 1, 
\hdots, T$ times and on $i = 1, \hdots, m$ users.
We assume that the reward at time $t$ is generated with a person and time specific
mean, 
\[E[Y_{i t}| \bX_{i t}, A_{i t}]= \bZ_{i t} \bbeta + \bZ^{\bu}_{i t} \bu_{i} +
\bZ^{\bv}_{i t} \bv_{t} \]
where $\bZ_{i t}= f(\bX_{i t}, A_{i t})$, $\bZ_{i t}^{\bu}=
f^{\bu}(\bX_{i t}, A_{i t})$ and $\bZ_{i t}^{\bv}= f^{\bv}(\bX_{i t}, A_{i t})$ are known
features of the context $\bX_{i t}$ and action $A_{i t}$. The $(\bbeta, \bu_i, \bv_t)$ are
unknown parameters; in particular $\bu_i$ is the vector of $i$th user parameters and
$\bv_{t}$ is the vector of time $t$ parameters.  Time $t$ corresponds to 
\lq\lq time-since-under-treatment\rq\rq\ for a user.
User-specific parameters, $\bu_{i}$, capture unobserved user variables that
influence the reward at all times $t$; in mHealth unobserved user variables may
include level of social support for activity, pre-existing problems or preferences that
make activity difficult. The time-specific parameters, $\bv_{t}$ capture
unobserved \lq\lq time-since-under-treatment\rq\rq\ variables that influence the reward
for all users. In mHealth unobserved \lq\lq time-since-under-treatment\rq\rq\
variables might include treatment fatigue, decreasing motivation, etc.

We use the following Bayesian mixed effects model 
\cite{laird1982random} for the reward $Y_{i t}$ as in \cite{tomkins2020intelligentpooling}:
\begin{equation}
\label{MainMod1}
\begin{array}{c}
  Y_{i t} | \bbeta, \bu_{i}, \bv_{t}, \sigsqeps \simind N(\bZ_{i t}\bbeta +
  \bZ_{it}^{\bu} \bu_{i} + \bZ_{it}^{v} \bv_{t}, \ \sigsqeps).
\end{array}
\end{equation}
We let $t$ represent the index for the current time since under treatment and $\tau$ be 
the index from the beginning of a study until the current time, i.e., 
$1 \le \tau \le t$. The algorithm is designed with independent Gaussian priors on 
the unknown parameters:
\begin{equation}
\label{MainMod2}
\begin{array}{c}
  \bbeta \sim N(\bmu_{\bbeta}, \bSigma_{\bbeta}), \quad
  \bu_{i} | \bSigma^{\bu} \simind N(\bzero, \bSigma^{\bu}), \ 1 \le i \le m, 
  \quad \bv_{\tau} | \bSigma^{\bv} \simind N(\bzero, \bSigma^{\bv}), \ 1\le \tau \le t.
\end{array}
\end{equation}
The $\bu_i$ and $\bv_{\tau}$ are called random effects in the
statistical literature and the model in (\ref{MainMod1}) and (\ref{MainMod2})
is often referred to as a linear mixed effects model \cite{laird1982random} or a linear mixed model with crossed random effects (e.g., \cite{baayen2008mixed, jeon2017variational}).
At each time, $t$, the TS Algorithm in Algorithm \ref{alg:TSalgo} 
is used to select the action, $A_{it}$, 
based on the context $\bX_{it}$. That is, we compute the posterior distribution 
for $\btheta_{it}$ where $\btheta_{it} = [\bbeta \hspace{1.5mm} \bu_{i} \hspace{1.5mm} \bv_{t}]^{T}$ 
and for context $\bX_{it}=\bx$, select treatment $A_{i t} = k$
with posterior probability
\begin{equation}
\label{eqn:randProb}
  \mbox{Pr}_{
  \btheta_{it} \sim N \left( \bmu_{p(\btheta_{it})}, \bSigma_{p(\btheta_{it})}
  \right)}\Bigg( E\left[Y_{it}| \bX_{it} = \bx, A_{it} = k
  \right] = \displaystyle{\max_{a = 1, \hdots, K}} \Big\{ E[Y_{it}| \bX_{it} =
  \bx, A_{it} = a] \Big\}  \Bigg)
\end{equation}
where $\left( \bmu_{\btheta_{it}}, \bSigma_{\btheta_{it}}\right)$ are  the posterior
mean and covariance matrix given in the sub-blocks of (\ref{eqn:FullPost}).

Further detail, including notation, on the Bayesian mixed effects model 
components is given in the Supplement in Section 
\ref{Sec:BayesMEM}. The form of the posterior updates is provided 
in Section \ref{Sec:PostUpdates}. The Na\"{i}ve EB procedure is 
explained in Sections \ref{Sec:PEB} and \ref{Sec:EM} and the 
Na\"{i}ve EB algorithm is provided in Algorithm \ref{alg:EMforEB} in 
Section \ref{Sec:NaiveEM}. 

\begin{minipage}{0.47\textwidth}
\begin{algorithm}[H]
    \centering
    \caption{\textit{\small{Physical activity suggestions.}}}\label{alg:TSalgo}
    \begin{algorithmic}[1] 
       \begin{small}
            \State{\textbf{Initialize:} $\hat{\bSigma}^{(0)}$}
            \For{$t \in \{ t_{1}, \hdots, t_{T} \}:$}
               \For{for $\tau = 1, \hdots, t:$}
               \State{Receive context features $\bX_{i\tau}$ for user $i$ and time $\tau$}
               \State{Obtain posterior $p(\btheta_{i \tau})$ using Result \ref{Res:Main}}
               \State{Compute randomization probability $\pi$ in (\ref{eqn:randProb})}
               \State{Sample treatment $A_{i\tau} \sim \operatorname{Bern} \left({\pi}\right)$}
               \State{ Observe reward $Y_{i \tau}$}
              \If{\textbf{$\tau = t:$}}
                  \State{Update $\hat{\bSigma}$ with Alg. \ref{alg:EMforEBStream}}
                  \State{Update posterior $p(\btheta_{it})$}
                \EndIf
               \EndFor
            \EndFor
        \end{small}
    \end{algorithmic}
\end{algorithm}
\end{minipage}
\begin{minipage}{0.47\textwidth}
  \begin{algorithm}[H]
      \centering
      \caption{\textit{\small{Streamlined EB for fitting.}}}\label{alg:EMforEBStream}
      \begin{algorithmic}[1] 
         \begin{small}
          \If{log-likelihood not converged}
            \State{\textbf{Initialize:} $\hat{\bSigma}^{(0)}$ ; \quad $\ell = 0$}
            \State{\textbf{Repeat:}}
            \State{\quad \textbf{E-step:} Compute components of $\bmu_{p(\btheta)}$ and} 
            \State{\quad sub-blocks of $\bSigma_{p(\btheta)}$:}
            \State{\quad $\Ssc\thickarrow 
            \TinySolveTwoLevelSparseLeastSquares(\{(\bveci,\Bmati,
            \Bmatdoti): 1\le i\le m \})$}
            \State{\quad \quad where $\Ssc$ returns $\bx_{1}$, $\bA^{11}$, 
              $\bx_{2, i}$, $\bA^{22, i}$ and}
            \State{\quad \quad $\bA^{12, i}, 1\le i \le m$.}
            \State{\quad \textbf{M-step:} Compute variance components in}
            \State{\quad $\hat{\bSigma}^{(\ell+1)}$ via equation (\ref{Eqn:EM_var_updates})} 
            \State{\quad $\ell \leftarrow \ell + 1$}
          \EndIf 
          \end{small}
      \end{algorithmic}
  \end{algorithm}
  \end{minipage}
\subsubsection{Computational Challenges}\label{Sec:CompCons}
At each iteration in Algorithm \ref{alg:EMforEB}, computation of
$p(\btheta)$ requires solving the sparse matrix linear system
$\bC^{\top} \bR^{-1} \bC + \bD = \bC^{\top} \bR^{-1} \bY + \bo$
%
\begin{figure}[h]
  \centering
  \subfigure{\includegraphics[width=0.15\textwidth]{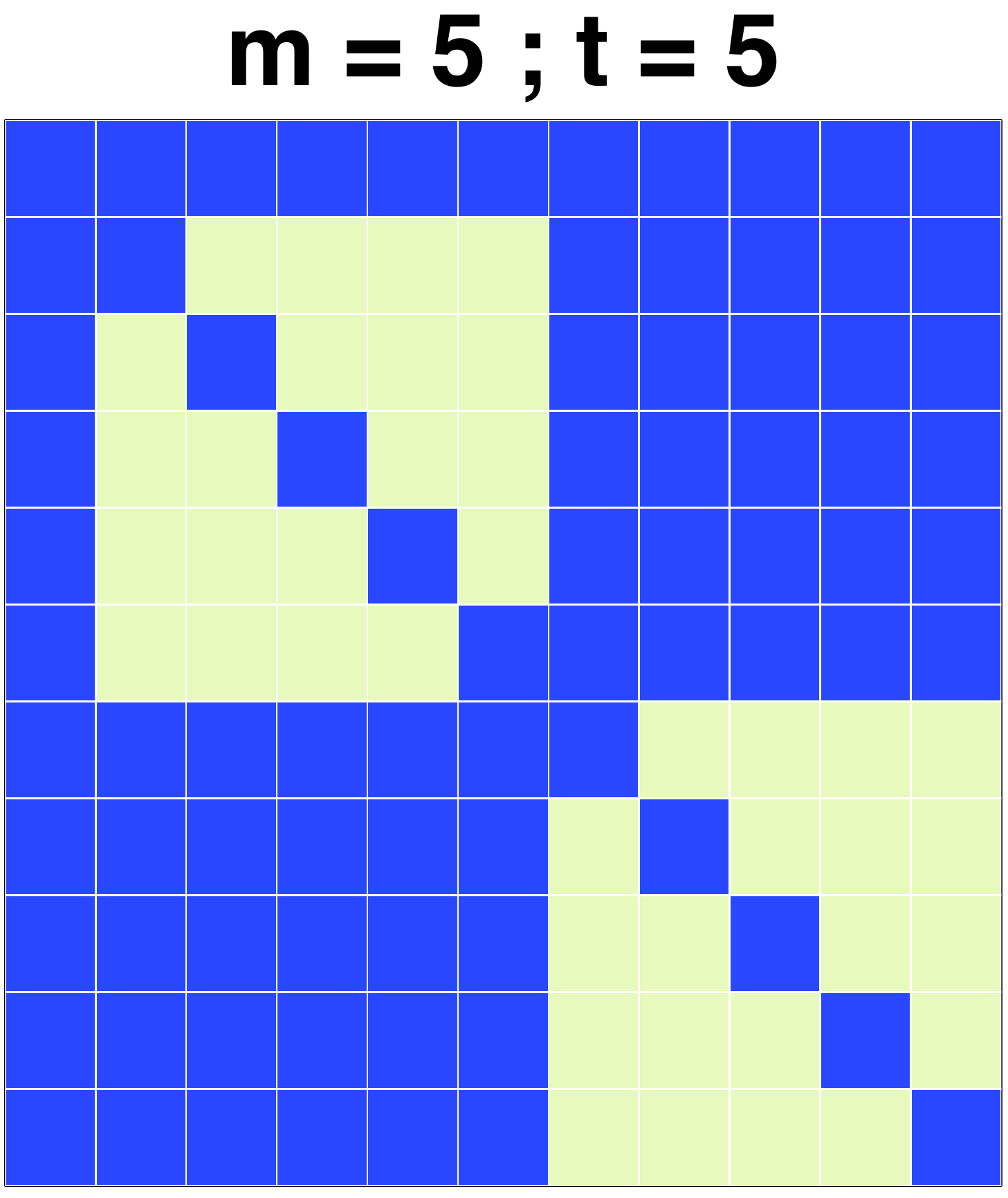}}
  \subfigure{\includegraphics[width=0.15\textwidth]{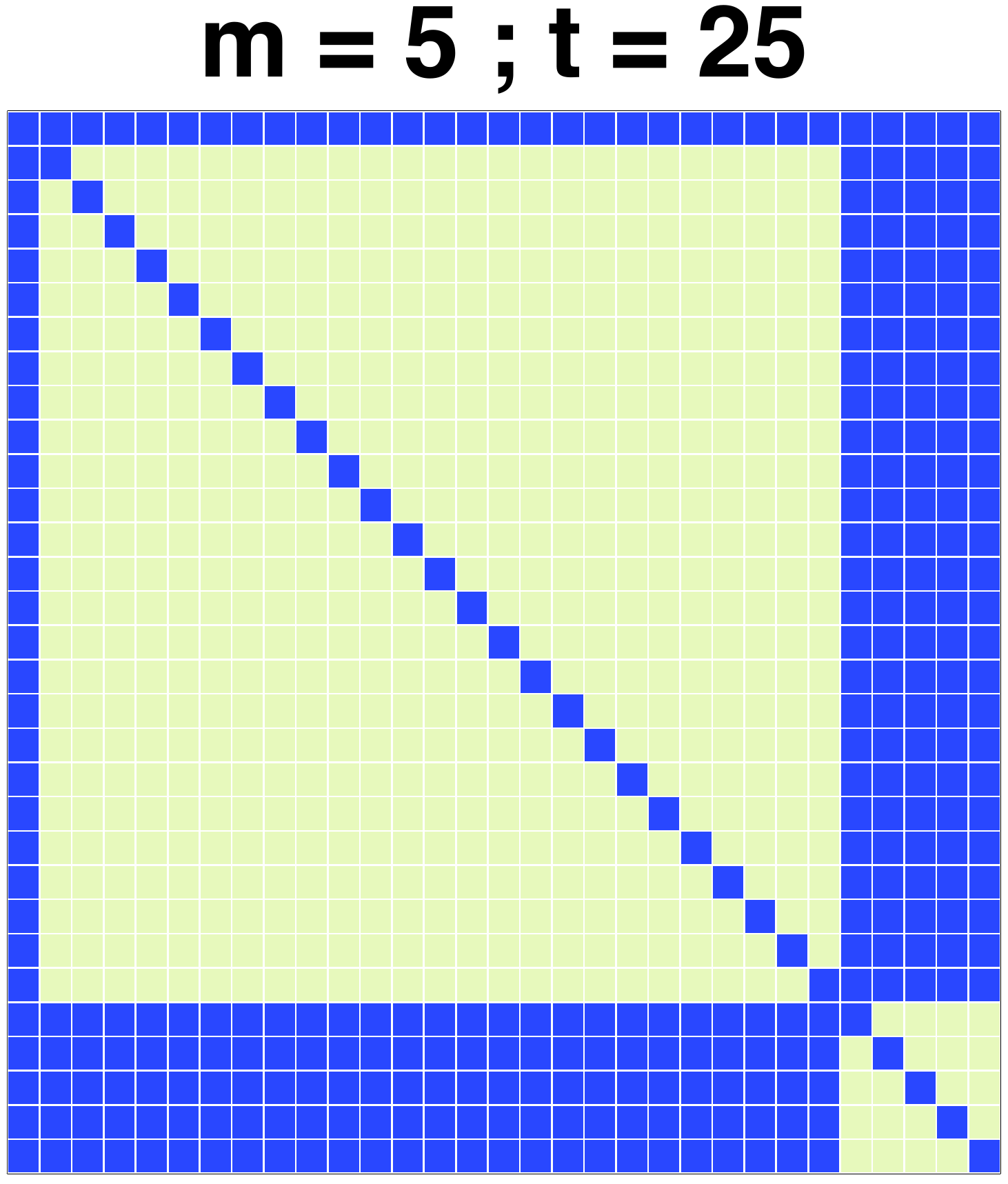}}
  \subfigure{\includegraphics[width=0.15\textwidth]{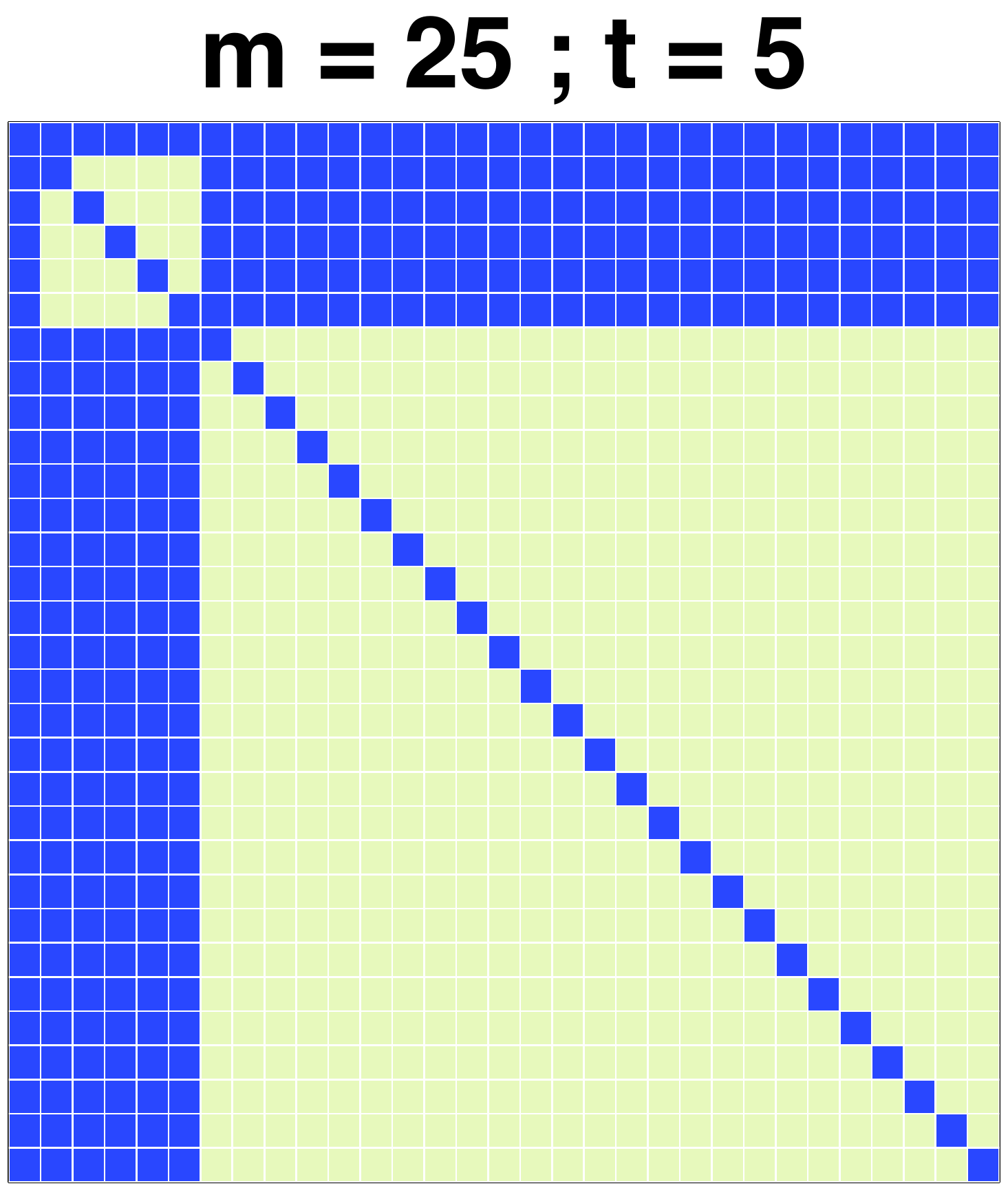}}
  \subfigure{\includegraphics[width=0.15\textwidth]{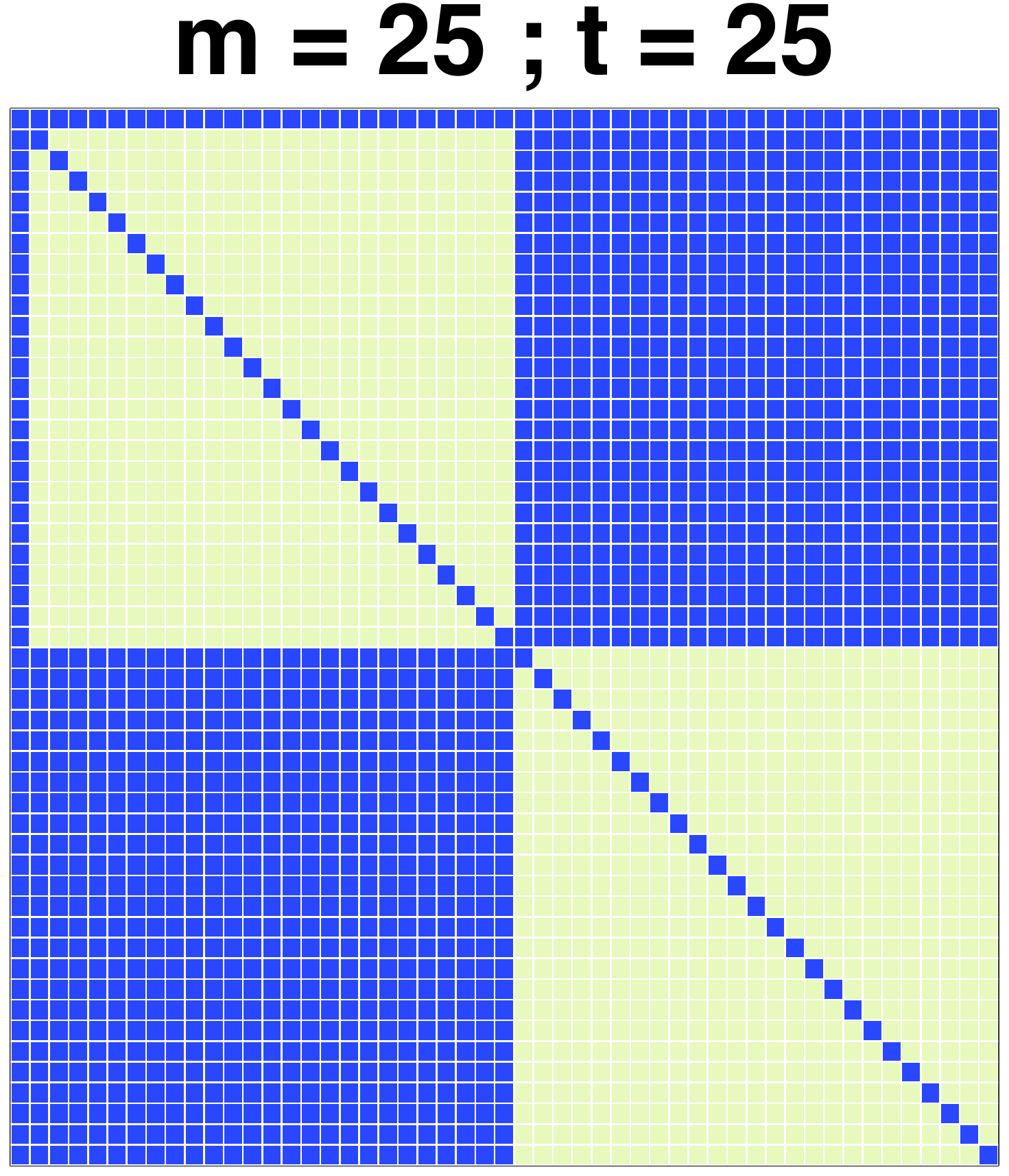}}
  \caption{\textit{\small{Sparsity in $\bC^{\top} \bR^{-1} \bC + \bD$
                   under the Bayesian mixed effects model as represented
                   in (\ref{MainMod1}) and (\ref{MainMod2}) where 
                   $p=q_{u}=q_{v}=1$. \emph{Blue-square:} Non-zero $1 \times 1$ entries; 
                   \emph{yellow-square:} zero $1 \times 1$ entries.}}}
   \label{fig:ModelIISparsity}
\end{figure}
%
where the LHS has sparse structure
imposed by the random effects as exemplified in Figure \ref{fig:ModelIISparsity}.
This matrix has dimension $(p+m q_{u} + t q_{v}) \times (p+ m q_{u} + t q_{v})$. 
Often, the number of fixed effects parameters $p$, random effects parameters 
per user $q_{u}$ and random effects parameters per time $q_{v}$ are small. 
Consequently, na\"{i}ve computation of $p(\btheta)$ is $\mathcal{O}((m+t)^{3})$, 
i.e., cubic dependence on the number of random effects group sizes $m$ and $t$.
To address this computational problem, we take advantage of the block-diagonal 
structure in Figure \ref{fig:ModelIISparsity} and the fact that the closed form 
updates of the variance components in (\ref{Eqn:EM_var_updates}) require 
computation of only certain sub-blocks of $\bSigma_{p(\btheta)}$ and not 
the entire matrix.
\subsection{Streamlined EB}\label{sec:streamPEB}
Streamlined updating of $\bmu_{p(\btheta)}$ and each of the sub-blocks of $\bSigma_{p(\btheta)}$
required for the E-step in Alg. \ref{alg:EMforEB} can be embedded within the class
of two-level sparse matrix problems as defined in \cite{nolan2019streamlined} and is
encapsulated in Result 1. Result 1 is analogous and mathematically identical to Result 2 in
\cite{menictas2019streamlinedCrossed}. The difference being that the authors in
\cite{menictas2019streamlinedCrossed} do not apply their methodologies to the mHealth
setting and use full variational Bayesian inference for fitting as opposed to our use of
EB.

The streamlined equivalent of Alg. \ref{alg:EMforEB} is given in Alg.
\ref{alg:EMforEBStream}. Alg. \ref{alg:EMforEBStream} makes use of the
\SolveTwoLevelSparseLeastSquares\ algorithm which was first presented in
\cite{nolan2019streamlined} but also provided in the appendix of this article.
The computing time and storage for the streamlined updating of $\bmu_{p(\btheta)}$ and each of the sub-blocks
of $\bSigma_{p(\btheta)}$ required for the E-step in Alg. \ref{alg:EMforEBStream}
becomes $\mathcal{O}(m t^3)$. For moderate sized $t$, this reduces to
$\mathcal{O}(m)$. 
\section{Performance Assessment and Comparison}\label{Sec:PerfAssComp}

We compare the timing and accuracy of our streamlined method against 
state of the art software, \texttt{GPyTorch} \cite{gardner2018gpytorch},
which is a highly efficient implementation of Gaussian Process
Regression modeling, with GPU acceleration. Note that the Bayesian linear mixed effects
model as given in (\ref{MainMod1}) and (\ref{MainMod2}) is equivalent to a
Gaussian Process regression model with a simple structured kernel. 
We use \SEB\ to refer to the streamlined EB algorithm, \CPU\ for 
EB fitting using \texttt{GPyTorch} with CPU and \GPU\ for 
EB fitting using \texttt{GPyTorch} with GPU. \SEB\ and \CPU\ 
were computed using an Intel Xeon CPU E5-2683. \GPU\ was computed using 
an Nvidea Tesla V100-PCIE-32GB.

\subsection{Batch Speed Assessment}\label{Sec:SpeedAss}

We simulated batch data according to versions of the Bayesian mixed effects 
model in (\ref{MainMod1}) and (\ref{MainMod2}) for a continuous predictor 
generated from the Uniform distribution on the unit interval. The true parameter 
values were set to

\begin{equation*}
    \begin{array}{c}
        \bbeta_{\mbox{\tiny true}}=\left[
        \begin{array}{c}
        0.58 \\
        1.98
        \end{array}
        \right],
        \quad
        \bSigma^{\bu}_{\mbox{\tiny true}}=
        \left[
        \begin{array}{cc}
        0.32 & 0.09\\
        0.09 & 0.42
        \end{array}
        \right],
        \quad
        \bSigma^{\bv}_{\mbox{\tiny true}}=
        \left[
        \begin{array}{cc}
        0.30 & 0\\
        0 & 0.25
        \end{array}
        \right], \mbox{ and} \quad
        {\sigsqeps}_{\mbox{\tiny, true}} = 0.3,
    \end{array}
\end{equation*}

and, $t$ was set to $30$, the number of data points for each user and 
time period, $n$, was set to 5. Four studies were run with differing 
values for the number of users $m \in \{ 10, 50, 100,10000 \}$. The 
total number of data points is then $ntm$, $\mbox{datapoints} \in 
\{ 1500, 7500, 15000, 1500000 \}$. We simulated 50 replications of 
the data for each $m$ and recorded computational times for 
from \CPU, \GPU\ and \SEB.
Alg. \ref{alg:EMforEBStream} was implemented in \texttt{R} with 
\texttt{Fortran 77} wrappers. EM iterations stopped once 
the absolute difference between successive expected complete 
data log-likelihood values fell below $10^{-5}$. The stopping 
criterion was the same for gPyTorch with the maximum number 
of iterations set to 15.

\begin{figure}
    \centering
    \begin{minipage}{.58\textwidth}
        \centering
        \includegraphics[width=\textwidth]{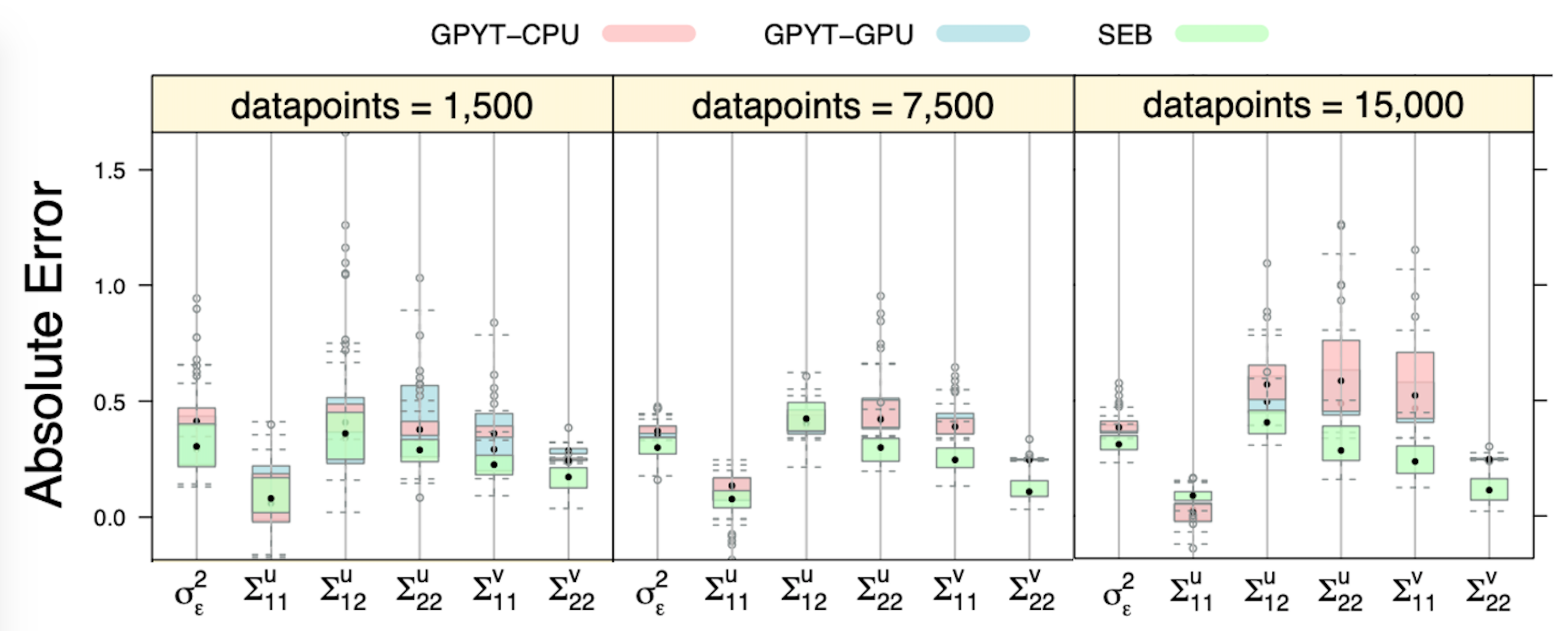}
        \caption{\small{\textit{Summary of the batch speed assessment. 
      Absolute error values for each variance component estimated
      using \SEB\, \CPU\ and \GPU\ summarized as a boxplot.}}}
        \label{fig:AbsErrII} 
    \end{minipage}%
    \hspace{2mm}
    \begin{minipage}{0.38\textwidth}
        \centering
       \includegraphics[width=\textwidth]{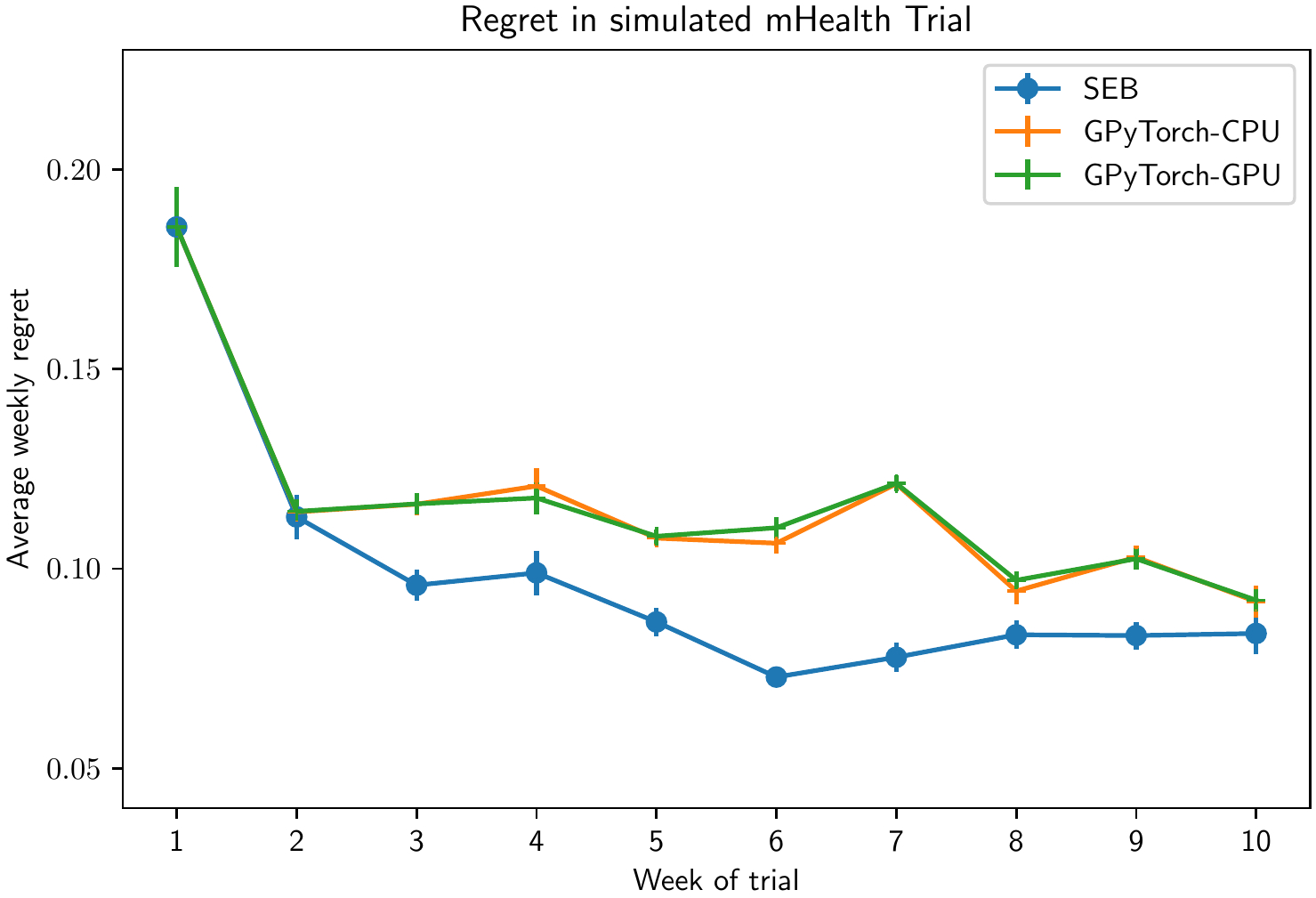}
       \caption{\small{\textit{Regret averaged across all users for each week in the
      simulated mHealth trial associated with approaches \SEB, \CPU\ and \GPU.}}}
       \label{fig:regretFig}
    \end{minipage}
\end{figure}

Table \ref{tab:TimesII} shows the mean (std dev) of elapsed
computing times in seconds for estimation of the variance components using
\SEB, \CPU\ and \GPU. NA values mean that computational burden was too high 
to produce results. Figure \ref{fig:AbsErrII} shows the absolute error
values for each variance components estimated using \SEB, \CPU\ and \GPU\
summarized as a boxplot.

\begin{table}[h]
\begin{center}
\begin{small}
\begin{tabular}{cccc}
        \hline\\[-2.3ex]
        Datapoints      & \SEB & \CPU & \GPU \\[0.2ex]
        \hline\\[-2.3ex]
        1,500     & 0.7 (0.10)    & 5.8 (0.14)    & 1.5 (0.16) \\[0.8ex]
        7,500     & 1.7 (0.15)    & 163.8 (1.81)  & 1.3 (0.04) \\[0.8ex]
        15,000    & 2.8 (0.21)    & 736.2 (38.36) & 5.2 (0.03) \\[0.8ex]
        1,500,000 & 322.1 (24.82) & NA (NA)       & NA (NA)    \\
        \hline
\end{tabular}
\end{small}
\end{center}
\caption{\small{\textit{Mean (std dev) of computing times in seconds for 
      estimation of the variance components using \SEB\ 
      via Alg. \ref{alg:EMforEBStream}, \CPU\ and \GPU\ for 
      comparison.}}
}
\label{tab:TimesII}
\end{table}

\subsection{Online TS Contextual Bandit mHealth
Simulation Study}\label{sec:onlineAlg}

Next, we evaluate our approach in a simulated mHealth study designed to 
capture many of the real-world difficulties of mHealth clinical trials. 
Users in this simulated study are sent interventions multiple times 
each day according to Alg. \ref{alg:TSalgo}. Each intervention 
represents a message promoting healthy behavior.
In this setting there are 32 users and each user is in the study 
for 10 weeks. Users join the study in a staggered fashion, such that 
each week new users might be joining or leaving the study. Each day 
in the study users can receive up to 5 mHealth interventions.

In Figure \ref{fig:regretFig} we show the ability of our streamlined
algorithm to minimize regret where real data is used to inform the simulation.
We also compare our approach to \CPU\ and \GPU. For all users we show
the average performance for their $n$th week in the study. For example,
we first show the average regret across all users in their first week
of the study, however this will not be the same calendar time week,
as users join in a staggered manner.
The average total time (standard deviation) for estimation of the
variance components was 757.1 (76.48) using \CPU, 7.5 (0.27) using
\GPU\ and 7.3 (0.16) using \SEB.

\section{Conclusion}

Inspection of the computational times in Table
\ref{tab:TimesII} shows \SEB\ achieving the lowest average
time across all simulations and data set sizes, compared to \CPU\ and \GPU. 
In the starkest case this results in a
98.96\% reduction in running time. Even in the more modest
comparison, \SEB\ timing is similar to that of \GPU\ but doesn't
require advanced hardware.




\begin{ack}
  This work was supported in part by U.S National Institutes of Health grants 
  R01AA23187 (NIH /NI AAA), P50DA039838 (NIH/NIDA) and U01CA229437 (NIH/NCI).
\end{ack}

\bibliographystyle{plain}
\bibliography{mybib}

\begin{thebibliography}{10}

\bibitem{ambikasaran2015fast}
Sivaram Ambikasaran, Daniel Foreman-Mackey, Leslie Greengard, David~W Hogg, and
  Michael O’Neil.
\newblock Fast direct methods for gaussian processes.
\newblock {\em IEEE transactions on pattern analysis and machine intelligence},
  38(2):252--265, 2015.

\bibitem{baayen2008mixed}
R~Harald Baayen, Douglas~J Davidson, and Douglas~M Bates.
\newblock Mixed-effects modeling with crossed random effects for subjects and
  items.
\newblock {\em Journal of memory and language}, 59(4):390--412, 2008.

\bibitem{berlin1990meta}
Jesse~A Berlin and Graham~A Colditz.
\newblock A meta-analysis of physical activity in the prevention of coronary
  heart disease.
\newblock {\em American journal of epidemiology}, 132(4):612--628, 1990.

\bibitem{billinger2014physical}
Sandra~A Billinger, Ross Arena, Julie Bernhardt, Janice~J Eng, Barry~A
  Franklin, Cheryl~Mortag Johnson, Marilyn MacKay-Lyons, Richard~F Macko,
  Gillian~E Mead, Elliot~J Roth, et~al.
\newblock Physical activity and exercise recommendations for stroke survivors:
  a statement for healthcare professionals from the american heart
  association/american stroke association.
\newblock {\em Stroke}, 45(8):2532--2553, 2014.

\bibitem{bogunovic2016time}
Ilija Bogunovic, Jonathan Scarlett, and Volkan Cevher.
\newblock Time-varying gaussian process bandit optimization.
\newblock In {\em Artificial Intelligence and Statistics}, pages 314--323,
  2016.

\bibitem{carpenter2017stan}
Bob Carpenter, Andrew Gelman, Matthew~D Hoffman, Daniel Lee, Ben Goodrich,
  Michael Betancourt, Marcus Brubaker, Jiqiang Guo, Peter Li, and Allen
  Riddell.
\newblock Stan: A probabilistic programming language.
\newblock {\em Journal of statistical software}, 76(1), 2017.

\bibitem{casella1985introduction}
George Casella.
\newblock An introduction to empirical bayes data analysis.
\newblock {\em The American Statistician}, 39(2):83--87, 1985.

\bibitem{chowdhury2017kernelized}
Sayak~Ray Chowdhury and Aditya Gopalan.
\newblock On kernelized multi-armed bandits.
\newblock In {\em Proceedings of the 34th International Conference on Machine
  Learning-Volume 70}, pages 844--853. JMLR. org, 2017.

\bibitem{dempster1977maximum}
Arthur~P Dempster, Nan~M Laird, and Donald~B Rubin.
\newblock Maximum likelihood from incomplete data via the em algorithm.
\newblock {\em Journal of the Royal Statistical Society: Series B
  (Methodological)}, 39(1):1--22, 1977.

\bibitem{desautels2014parallelizing}
Thomas Desautels, Andreas Krause, and Joel~W Burdick.
\newblock Parallelizing exploration-exploitation tradeoffs in gaussian process
  bandit optimization.
\newblock {\em The Journal of Machine Learning Research}, 15(1):3873--3923,
  2014.

\bibitem{djolonga2013high}
Josip Djolonga, Andreas Krause, and Volkan Cevher.
\newblock High-dimensional gaussian process bandits.
\newblock In {\em Advances in Neural Information Processing Systems}, pages
  1025--1033, 2013.

\bibitem{dorie2015blme}
V~Dorie.
\newblock blme: Bayesian linear mixed-effects models.
\newblock {\em URL: https://CRAN. R-project. org/package= blme R package
  version}, pages 1--0, 2015.

\bibitem{gardner2018gpytorch}
Jacob Gardner, Geoff Pleiss, Kilian~Q Weinberger, David Bindel, and Andrew~G
  Wilson.
\newblock Gpytorch: Blackbox matrix-matrix gaussian process inference with gpu
  acceleration.
\newblock In {\em Advances in Neural Information Processing Systems}, 2018.

\bibitem{hoffman2011portfolio}
Matthew~D Hoffman, Eric Brochu, and Nando de~Freitas.
\newblock Portfolio allocation for bayesian optimization.
\newblock In {\em UAI}, pages 327--336. Citeseer, 2011.

\bibitem{jeon2017variational}
Minjeong Jeon, Frank Rijmen, and Sophia Rabe-Hesketh.
\newblock A variational maximization--maximization algorithm for generalized
  linear mixed models with crossed random effects.
\newblock {\em psychometrika}, 82(3):693--716, 2017.

\bibitem{klasnja2019efficacy}
Predrag Klasnja, Shawna Smith, Nicholas~J Seewald, Andy Lee, Kelly Hall, Brook
  Luers, Eric~B Hekler, and Susan~A Murphy.
\newblock Efficacy of contextually tailored suggestions for physical activity:
  A micro-randomized optimization trial of heartsteps.
\newblock {\em Annals of Behavioral Medicine}, 53(6):573--582, 2019.

\bibitem{krause2011contextual}
Andreas Krause and Cheng~S Ong.
\newblock Contextual gaussian process bandit optimization.
\newblock In {\em Advances in neural information processing systems}, pages
  2447--2455, 2011.

\bibitem{laird1982random}
Nan~M Laird, James~H Ware, et~al.
\newblock Random-effects models for longitudinal data.
\newblock {\em Biometrics}, 38(4):963--974, 1982.

\bibitem{li2010contextual}
Lihong Li, Wei Chu, John Langford, and Robert~E Schapire.
\newblock A contextual-bandit approach to personalized news article
  recommendation.
\newblock In {\em Proceedings of the 19th international conference on World
  wide web}, pages 661--670. ACM, 2010.

\bibitem{liu2018gaussian}
Haitao Liu, Yew-Soon Ong, Xiaobo Shen, and Jianfei Cai.
\newblock When gaussian process meets big data: A review of scalable gps.
\newblock {\em arXiv preprint arXiv:1807.01065}, 2018.

\bibitem{menictas2019streamlinedCrossed}
Marianne Menictas, Gioia Di~Credico, and Matt~P Wand.
\newblock Streamlined variational inference for linear mixed models with
  crossed random effects.
\newblock {\em arXiv preprint arXiv:1910.01799}, 2019.

\bibitem{morris1983parametric}
Carl~N Morris.
\newblock Parametric empirical bayes inference: theory and applications.
\newblock {\em Journal of the American statistical Association},
  78(381):47--55, 1983.

\bibitem{nolan2019streamlined}
Tui~H Nolan, Marianne Menictas, and Matt~P Wand.
\newblock Streamlined computing for variational inference with higher level
  random effects.
\newblock {\em arXiv preprint arXiv:1903.06616}, 2019.

\bibitem{nolan2019solutions}
Tui~H Nolan and Matt~P Wand.
\newblock Solutions to sparse multilevel matrix problems.
\newblock {\em arXiv preprint arXiv:1903.03089}, 2019.

\bibitem{srinivas2009gaussian}
Niranjan Srinivas, Andreas Krause, Sham~M Kakade, and Matthias Seeger.
\newblock Gaussian process optimization in the bandit setting: No regret and
  experimental design.
\newblock {\em arXiv preprint arXiv:0912.3995}, 2009.

\bibitem{tomkins2020intelligentpooling}
Sabina Tomkins, Peng Liao, Predrag Klasnja, and Susan Murphy.
\newblock Intelligentpooling: Practical thompson sampling for mhealth.
\newblock {\em arXiv preprint arXiv:2008.01571}, 2020.

\bibitem{wang2016optimization}
Zi~Wang, Bolei Zhou, and Stefanie Jegelka.
\newblock Optimization as estimation with gaussian processes in bandit
  settings.
\newblock In {\em Artificial Intelligence and Statistics}, pages 1022--1031,
  2016.

\end{thebibliography}

\newpage

\renewcommand{\theequation}{S.\arabic{equation}}
\renewcommand{\thesection}{S.\arabic{section}}
\renewcommand{\thetable}{S.\arabic{table}}
\renewcommand{\thealgorithm}{S.\arabic{algorithm}}
\renewcommand{\theresult}{S.\arabic{result}}
\setcounter{equation}{0}
\setcounter{table}{0}
\setcounter{section}{0}
\setcounter{page}{1}
\setcounter{footnote}{0}
\setcounter{algorithm}{0}
\setcounter{result}{0}

\null
\vskip5mm
\centerline{\Large \bf Supplementary Material}
\vskip3mm

\centerline{\Large Fast Physical Activity Suggestions:}
\centerline{\Large Efficient Hyperparameter Learning in Mobile Health}
\vskip6mm

\section{Bayesian Mixed Effects Model Components}\label{Sec:BayesMEM}
We define the following data matrices
\begin{equation*}
  \begin{array}{c}
    \bY \equiv \displaystyle{\stack{1\le i\le m}} \left( \bY_{i} \right), 
    \hspace{2mm}
    \bY_{i} \equiv \displaystyle{\stack{1\le \tau \le t}} \left( Y_{i\tau} \right), 
    \hspace{2mm}
    \bZ \equiv \displaystyle{\stack{1\le i \le m}} \left( \bZ_{i} \right) ,
    \hspace{2mm}
    \bZ_{i} \equiv \displaystyle{\stack{1\le \tau \le t}} \left( \bZ_{i\tau} \right),
    \\[2ex]
    \bZ^{\bu}_{i} \equiv \displaystyle{\stack{1 \le \tau \le t}} \left( \bZ^{\bu}_{i\tau} \right),
      \hspace{2mm}
      \bZ^{\bv}_{i} \equiv \displaystyle{\blockdiag{1 \le \tau \le t}} \left( 
       \bZ^{\bv}_{i\tau} \right),
      \hspace{2mm}
      \bZ^{\bu \bv} \equiv \left[ \displaystyle{\blockdiag{1\le i\le m}} 
      \left( \bZ^{\bu}_{i} \right) \hspace{1.5mm} \displaystyle{\stack{1\le i\le m}} \left( 
      \bZ^{\bv}_{i} \right) \right],
    \end{array}
\end{equation*}
and the parameter vectors $\bbeta \equiv \left[ \beta_{0} \ \hdots \ 
\beta_{p-1} \right]^{\top}$, $\bu \equiv \left[ \bu_{1} \ \hdots \ \bu_{m} \right]^{\top}$, 
$\bv \equiv \left[ \bv_{1} \ \hdots \ \bv_{t} \right]^{\top}$, where, 
as before, $\bZ^{\bu}_{it} = f^{\bu}(\bX_{it}, A_{it})$ and
$\bZ^{\bv}_{it} = f^{\bv}(\bX_{it}, A_{it})$ represent the known features
of the context $\bX_{it}$ and action $A_{it}$.
The dimensions of matrices, for $1 \le i \le m$ and $1 \le \tau \le t$, are: 
$\bZ_{i\tau}\ \mbox{is $1 \times p$}$, $\bbeta\ \mbox{is $p\times1$}$, 
$\bZ^{\bu}_{i\tau} \ \mbox{is $1 \times q_{u}$}$, $\bZ^{\bv}_{i\tau}\ \mbox{is $1 \times q_{v}$}$, 
$\bu_i\ \mbox{is $q_{u}\times 1$}$, $\bv_{\tau}\ \mbox{is $q_{v}\times 1$}$, 
$\bSigma_{\bu}\ \ \mbox{is $q_{u}\times q_{u}$}$ and 
$\bSigma_{\bv}\ \ \mbox{is $q_{v}\times q_{v}$}$.
%
%

\section{Posterior Updates}\label{Sec:PostUpdates}
The posterior distribution $\btheta_{it}$
for the immediate treatment effect for user $i$ at time $t$ is updated
and then used to assign treatment in the subsequent time point,
$t + 1$. Here, we show the form of
the full posterior for $\btheta = \left[ \bbeta \ \bu \ \bv \right]^{\top}$.
Define
\begin{equation*}
  \begin{array}{c}
    \bC \equiv \left[ \bZ \, \bZ^{\bu \bv} \right], \quad
    \bD \equiv \left[
    \begin{array}{ccc}
       \bSigma_{\bbeta}^{-1} & \bzero & \bzero \\
       \bzero & \bI_{m} \otimes \hat{\bSigma}_{\bu}^{-1} & \bzero \\
       \bzero & \bzero & \bI_{t} \otimes \hat{\bSigma}_{\bv}^{-1}
    \end{array} \right], \quad 
    \bR \equiv \hat{\sigsqeps} \bI, \quad \mbox{and} \quad
    \bo \equiv \left[
    \begin{array}{c}
       \bSigma_{\bbeta}^{-1} \bmu_{\bbeta} \\
       \bzero
    \end{array} \right].
\end{array}
\end{equation*}
The estimated posterior distribution for the fixed and random reward
effects vector $\btheta$ is
\begin{equation}
  \label{eqn:FullPost}
  \begin{array}{c}
    \btheta \, | \, \hat{\bSigma} \sim N \left(
    \bmu_{p(\btheta)}, \, \bSigma_{p(\btheta)} \right), \quad 
    \mbox{where} \quad 
    \hat{\bSigma} \equiv (\hat{\sigsqeps}, \hat{\bSigma}_{\bu}, 
    \hat{\bSigma}_{\bv}),
    \\[1ex]
    \bSigma_{p(\btheta)} = \left(\bC^{\top} \bR^{-1} \bC + \bD \right)^{-1},
    \quad \mbox{and}
    \quad
    \bmu_{p(\btheta)} = (\bC^{\top} \bR^{-1} \bC + \bD )^{-1}
    (\bC^{\top} \bR^{-1} \by + \bo).
  \end{array}
\end{equation}
The focus of this work is to enable fast incremental estimation of the variance
components $\bSigma \equiv (\sigsqeps, \bSigma_{\bu}, \bSigma_{\bv})$.
We describe a natural, but computationally challenging approach for estimating
these variances in Section \ref{Sec:PEB} and our
streamlined alternative approach in Section \ref{sec:streamPEB}.
\subsection{Parametric Empirical Bayes}\label{Sec:PEB}
At each time, $t$, the EB \cite{morris1983parametric, casella1985introduction} 
procedure maximizes the  marginal likelihood based on
all user data up to and including data at time $t$ with respect to $\bSigma$.
The marginal likelihood of $\bY$ is $\bY \, | \, \bSigma \sim N (\bzero, \, \bC \bD 
\bC^{\top} + \sigsqeps \bI)$ and has the following form
\begin{equation*}
  \begin{array}{l}
     p(\bY \, | \, \bSigma) = (2 \pi)^{-\smhalf mt}
     |\bC \bD \bC^{\top} + \sigsqeps \bI|^{-\smhalf}
     \exp \left\{ -\smhalf \bY^{\top}
     \left( \bC \bD \bC^{\top} + \sigsqeps \bI \right)^{-1} \bY \right\}.
  \end{array}
\end{equation*}
The maximization is commonly done via  the Expectation Maximisation (EM) algorithm
\cite{dempster1977maximum}.
\subsection{Expectation Maximization (EM) Method}\label{Sec:EM}
The {expected} complete data log likelihood is given by $L(\bSigma) = 
E\left[ \log p(\bY | \btheta, \bSigma) + \log p(\btheta | \bSigma) \right]$ 
where the expectation is over the distribution of $\btheta=\left[ \bbeta
\ \bu \ \bv \right]^{\top}$ given in (\ref{MainMod2}). The M-step yields
the following closed form $(\ell + 1)$ iteration estimates for the variance
components in $\hat{\bSigma}^{(\ell+1)}$:
\begin{equation}
  \label{Eqn:EM_var_updates}
  \begin{array}{l}
    \left(\hat{\sigma}^{2}_{\varepsilon}\right)^{(\ell + 1)} =
    \displaystyle{\sum_{i=1}^{m}} \displaystyle{\sum_{\tau=1}^{t}} \left\{ ||Y_{i\tau}
    - \bZ_{i\tau} \bmu_{p(\bbeta)} - \bZ^{\bu}_{i\tau} \bmu_{p(\bu_{i})}
    - \bZ^{\bv}_{i\tau} \bmu_{p(\bv_{\tau})} ||^{2} + \mbox{tr} \left(
    \bZ_{i\tau}^{\top} \bZ_{i\tau} \bSigma_{p(\bbeta)} \right) \right.
    \\[3ex]
    \quad\quad\quad\quad\quad\quad
    \left. + \ \mbox{tr} \left( {\bZ^{\bu}_{i\tau}}^{\top} \bZ^{\bu}_{i\tau}
    \bSigma_{p(\bu_{i})} \right) + \ \mbox{tr} \left( {\bZ^{\bv}_{i\tau}}^{\top}
    \bZ^{\bv}_{i\tau} \bSigma_{p(\bv_{\tau})} \right) 
    + \ \mbox{tr} \left( \bZ_{i\tau}^{\top} \bZ^{\bu}_{i\tau}
    \bCov_{p(\bbeta, \bu_{i})} \right)\right.
    \\[2ex]
    \quad\quad\quad\quad\quad\quad
    \left.  + \ \mbox{tr} \left( \bZ_{i\tau}^{\top}
    \bZ^{\bv}_{i\tau} \bCov_{p(\bbeta, \bv_{\tau})} \right) 
    + \ \mbox{tr} \left( {\bZ^{\bu}_{i\tau}}^{\top}
    \bZ^{\bv}_{i\tau} \bCov_{p(\bu_{i}, \bv_{\tau})} \right) \right\},
    \\[2ex]
    \hat{\bSigma}_{\bu}^{(\ell + 1)} = \frac{1}{m} \displaystyle{\sum_{i=1}^{m}} \left\{
    \bmu_{p(\bu_{i})} \bmu_{p(\bu_{i})}^{\top} + \bSigma_{p(\bu_{i})} \right\}, 
    \quad \quad 
    \hat{\bSigma}_{\bv}^{(\ell + 1)} = \frac{1}{t} \displaystyle{\sum_{\tau=1}^{t}} \left\{
    \bmu_{p(\bv_{\tau})} \bmu_{p(\bv_{\tau})}^{\top} + \bSigma_{p(\bv_{\tau})} \right\}.
  \end{array}
\end{equation}
where the posterior mean reward components for the fixed and random effects
\begin{equation}
  \label{eqn:betasubentries}
  \begin{array}{c}
      \bmu_{p(\bbeta)}, \ \bmu_{p(\bu_{i})}, \ 1 \le i \le m, \quad
      \bmu_{p(\bv_{\tau})}, \ 1 \le \tau \le t,
  \end{array}
\end{equation}
and the posterior variance-covariance reward components for the fixed and
random effects
\begin{equation}
  \label{eqn:sigmasubblocks}
  \begin{array}{c}
    \bSigma_{p(\bbeta)}, \ \bSigma_{p(\bu_{i})} \ \bSigma_{p(\bv_{\tau})},
    \bCov_{p(\bbeta, \bu_{i})}, \ \bCov_{p(\bbeta, \bv_{\tau})}, \
    \bCov_{p(\bu_{i}, \bv_{\tau})}, \quad 1 \le i \le m, \ 1 \le \tau \le t,
  \end{array}
\end{equation}
are computed in the E-step using equation (\ref{eqn:FullPost}). Note that
(\ref{eqn:betasubentries}) are the sub vectors in the the posterior mean $\bmu_{p(\btheta)}$
and (\ref{eqn:sigmasubblocks}) are sub-block entries in the posterior variance
covariance matrix $\bSigma_{p(\btheta)}$. The na\"{i}ve EM algorithm is given in
Alg. \ref{alg:EMforEB}. The challenge in Alg. \ref{alg:EMforEB} is computation of the
posterior at each iteration. 

\section{The Na\"{i}ve Expectation Maximization Algorithm}\label{Sec:NaiveEM}

\begin{algorithm}[h]
  \begin{center}
    \begin{minipage}[t]{150mm}
    \begin{small}
      \textbf{Initialize:} $\hat{\bSigma}^{(0)}$ \\[1ex]
      Set $\ell = 0$ \\[1ex]
      \textbf{repeat}
      \begin{itemize}
         \item[] \textbf{E-step:} Compute $\bmu_{p(\btheta)}$ and
                 $\bSigma_{p(\btheta)}$ via equation (\ref{eqn:FullPost})
                 to obtain necessary mean and covariance 
         \item[] components needed for the M-step.
         \item[] \textbf{M-step:} Compute variance components in    
                 $\hat{\bSigma}^{(\ell+1)}$
                 via equation (\ref{Eqn:EM_var_updates}).
         \item[] $\ell \leftarrow \ell + 1$
      \end{itemize}
      \textbf{until} log-likelihood converges
    \end{small}
    \end{minipage}
  \end{center}
  \caption{\textit{Na\"{i}ve EM Algorithm for EB estimates
  of the variance components in the Bayesian mixed effects model as given in (\ref{MainMod1}) and (\ref{MainMod2}).}}
  \label{alg:EMforEB}
\end{algorithm}
%

\section{Main Result}\label{sec:Res1}

\begin{result}\label{Res:Main}
{\textit{(Analogous to Result 2 in
\cite{menictas2019streamlinedCrossed}).}} \textit{
The posterior updates under the Bayesian mixed effects model as given in
(\ref{MainMod1}) and (\ref{MainMod2}) for $\bmu_{p(\btheta)}$
and each of the sub-blocks of $\bSigma_{p(\btheta)}$ are expressible as
a two-level sparse matrix least squares problem of the form $|| \bb - \bB \
\bmu_{p(\btheta)} ||^{2}$ where $\bb$ and the non-zero sub-blocks of $\bB$,
according to the notation in the appendix, are, for $1 \le i \le m$,}
$$
  \bb_{i} \equiv \left[
    \begin{array}{c}
      \sigma_{\varepsilon}^{-1} \bY_{i} \\[1ex]
      m^{-\smhalf} \bSigma_{\bbeta}^{-\smhalf} \bmu_{\bbeta} \\[1ex]
      \bzero \\[1ex]
      \bzero
    \end{array} \right],
  \quad
  \bBdot_i \equiv \left[
    \begin{array}{c}
      \sigma_{\varepsilon}^{-1} \bZ^{\bu}_{i} \\[1ex]
      \bO \\[1ex]
      \bO \\[1ex]
      \bSigma_{u}^{-\smhalf}
  \end{array} \right],
  \quad
  \bB_{i} \equiv \left[
    \begin{array}{cc}
      \sigma_{\varepsilon}^{-1} \bX_{i} & \sigma_{\varepsilon}^{-1} \bZ^{\bv}_{i} \\[1ex]
      m^{-\smhalf} \bSigma_{\bbeta}^{-\smhalf} & \bO  \\[1ex]
      \bO & m^{-\smhalf} \left( \bI_{t} \otimes \bSigma_{\bv}^{-\smhalf} \right) \\[1ex]
      \bO & \bO
  \end{array} \right],
$$
\textit{with each of these matrices having $\nadj = t + p + t q_{v} + q_{u}$
rows. The solutions are
$$
  \bmu_{p(\bbeta)} = \mbox{ first p rows of } \bx_{1}, \quad
  \bSigma_{p(\bbeta)} = \mbox{ top left } p \times p \mbox{ sub-block of } \bA^{11},
$$
\begin{equation*}
  \begin{array}{l}
    \displaystyle{\stack{1\le i\le m}} \left( \bmu_{p(\bu_{i})} \right)
     = \mbox{subsequent $q_{u}\times 1$ entries of}\ \bx_{1}
     \mbox{following $\bmu_{p(\bbeta)}$},
  \end{array}
\end{equation*}
\begin{equation*}
  \begin{array}{l}
    \bSigma_{p(\bu_{i})} =\mbox{ subsequent $q_{u} \times q_{u}$ diagonal
    sub-blocks} \mbox{of $\AUoo$ following $\bSigma_{p(\bbeta)}$},
  \end{array}
\end{equation*}
\begin{equation*}
  \begin{array}{l}
    \bCov_{p(\bbeta, \bu_{i})} = \mbox{ subsequent $p \times q^{\prime}$
    sub-blocks of $\AUoo$} \mbox{to the right of $\bSigma_{p(\bbeta)}$},
        \ \ 1\le i \le m,
  \end{array}
\end{equation*}
\begin{equation*}
  \begin{array}{l}
    \bmu_{p(\bv_{\tau})}= \bx_{2,\tau}, \ \ \bSigma_{p(\bv_{\tau})} = \bA^{22,\tau}, \quad
    \bCov_{p(\bbeta, \bv_{\tau})} = \mbox{first $p$ rows of $\bA^{12,\tau}$}
  \end{array}
\end{equation*}
and
\begin{equation*}
  \begin{array}{l}
    {\displaystyle\stack{1\le i\le m}}
    \Big( \bCov_{p(\bu_{i}, \bv_{\tau})} \Big) =
    \mbox{ remaining $q_{u}$rows of $\bA^{12,\tau}$,}
  \end{array}
\end{equation*}
$1 \le \tau \le t$, where the $\bx_{1}$, $\bx_{2,\tau}$, $\bA^{11}$,
$\bA^{22, \tau}$ and $\bA^{12, \tau}$ notation is given in the appendix.}
\end{result}
%

\section{The \LargerSolveTwoLevelSparseLeastSquares\ Algorithm}\label{sec:STLSLS}

The \SolveTwoLevelSparseLeastSquares\ algorithm is listed in \cite{nolan2019solutions} and
based on Theorem 2 of \cite{nolan2019solutions}. Given its centrality to Algorithm \ref{alg:EMforEBStream}
we list it again here. The algorithm solves a sparse version of the the least squares problem:
$$\min_{\bx}\Vert\bb-\bB\bx\Vert^2$$
which has solution $\bx=\bA^{-1}\bB^T\bb$ where $\bA=\bB^T\bB$ and
where $\bB$ and $\bb$ have the following structure:
\begin{equation}
\bB\equiv
\left[
\arraycolsep=2.2pt\def\arraystretch{1.6}
\begin{array}{c|c|c|c|c}
\Bmato &\Bmatdoto &\bO &\cdots&\bO\\
\hline
\Bmatt &\bO &\Bmatdott&\cdots&\bO\\
\hline
\vdots &\vdots &\vdots &\ddots&\vdots\\
\hline
\Bmatm &\bO &\bO &\cdots &\Bmatdotm
\end{array}
\right]
\quad\mbox{and}\quad
\bb=\left[
\arraycolsep=2.2pt\def\arraystretch{1.6}
\begin{array}{c}
\bveco \\
\hline
\bvect \\
\hline
\vdots \\
\hline
\bvecm \\
\end{array}
\right].
\label{eq:BandbFormsReprise}
\end{equation}
The sub-vectors of $\bx$ and the sub-matrices of $\AtLev$ corresponding to its
non-zero blocks of are labelled as follows:
\begin{equation}
\begin{array}{c}
\bx =
\left[
\arraycolsep=2.2pt\def\arraystretch{1.6}
\begin{array}{c}
\bx_1\\
\hline
\bx_{2,1}\\
\hline
\bx_{2,2}\\
\hline
\vdots\\
\hline
\bx_{2,m}
\end{array}
\right] \quad \mbox{and} \quad 
\AtLev^{-1}=
\left[
\arraycolsep=2.2pt\def\arraystretch{1.6}
\begin{array}{c|c|c|c|c}
\AUoo & \AUotCo & \AUotCt & \cdots  &\AUotCm \\
\hline
\AUotCoT & \AUttCo & \bigX & \cdots & \bigX \\
\hline
\AUotCtT & \bigX & \AUttCt & \cdots & \bigX \\
\hline
\vdots & \vdots  & \vdots & \ddots & \vdots \\
\hline
\AUotCmT & \bigX & \bigX & \cdots &\AUttCm \\
\end{array}
\right]
\label{eq:AtLevInv}
\end{array}
\end{equation}
with $\bigX$ denoting sub-blocks that are not of interest.
The \SolveTwoLevelSparseLeastSquares\ algorithm is given
in Algorithm \ref{alg:SolveTwoLevelSparseLeastSquares}.

\begin{algorithm}[h]
\begin{center}
\begin{minipage}[t]{150mm}
\begin{small}
\begin{itemize}
\setlength\itemsep{4pt}
\item[] Inputs: $\big\{\big(\bveci(\nadj_i\times1),
\ \Bmati(\nadj_i\times p),\ \Bmatdoti(\nadj_i\times q)\big): \ 1\le i\le m\big\}$
\item[] $\bomega_3\thickarrow\mbox{NULL}$\ \ \ ;\ \ \ $\bOmega_4\thickarrow\mbox{NULL}$
\item[] For $i=1,\ldots,m$:
\begin{itemize}
\setlength\itemsep{4pt}
\item[] Decompose $\Bmatdoti=\bQ_i\left[\begin{array}{c}
\bR_i\\
\bzero
\end{array}
\right]$ such that $\bQ_i^{-1}=\bQ_i^T$ and $\bR_i$ is upper-triangular.
\item[] $\cveczi\thickarrow\bQ_i^T\bveci\ \ \ ;\ \ \ \Cmatzi\thickarrow\bQ_i^T\Bmati$
        \ \ \ ; \ \ \ $\cvecoi\thickarrow\mbox{first $q$ rows of}\ \cveczi$
\item[] $\cvecti\thickarrow\mbox{remaining rows of}\ \cveczi$\ \ ;\ \
$\bomega_3\thickarrow
\left[
\begin{array}{c}
\bomega_3\\
\cvecti
\end{array}
\right]$
\item[]$\Cmatoi\thickarrow\mbox{first $q$ rows of}\ \Cmatzi$ \ \ \ ; \ \ \
       $\Cmatti\thickarrow\mbox{remaining rows of}\ \Cmatzi$ \ \ \ ; \ \ \
       $\bOmega_4\thickarrow
        \left[
        \begin{array}{c}
        \bOmega_4\\
        \Cmatti
        \end{array}
        \right]$
\end{itemize}
\item[] Decompose $\OmegaAtwoTwo=\bQ\left[\begin{array}{c}
\bR\\
\bzero
\end{array}
\right]$ such that $\bQ^{-1}=\bQ^T$ and $\bR$ is upper-triangular.
\item[] $\bc\thickarrow\mbox{first $p$ rows of $\bQ^T\bomega_3$}$
\ \ ; \ \ $\xveco\thickarrow\bR^{-1}\bc$ \ \ ; \ \
$\AUoo\thickarrow\bR^{-1}\bR^{-T}$
\item[] For $i=1,\ldots,m$:
\begin{itemize}
\setlength\itemsep{4pt}
\item[] $\xvectCi\thickarrow\bR_i^{-1}(\bc_{1i}-\Cmatoi\xveco)$ \ \ ; \ \
        $\AUotCi\thickarrow\,-\AUoo(\bR_i^{-1}\Cmatoi)^T$
\item[] $\AUttCi\thickarrow\bR_i^{-1}(\bR_i^{-T} - \Cmatoi\AUotCi)$
\end{itemize}
\item[] Output: $\Big(\xveco,\AUoo,\big\{\big(\xvectCi,\AUttCi,\AUotCi):\ 1\le i\le m\big\}\Big)$
\end{itemize}
\end{small}
\end{minipage}
\end{center}
\caption{\SolveTwoLevelSparseLeastSquares\ \textit{for solving the two-level sparse matrix
least squares problem: minimise $\Vert\bb-\bB\,\bx\Vert^2$ in $\bx$ and sub-blocks of $\bA^{-1}$
corresponding to the non-zero sub-blocks of $\bA=\bB^T\bB$. The sub-block notation is
given by (\ref{eq:BandbFormsReprise}) and (\ref{eq:AtLevInv}).}}
\label{alg:SolveTwoLevelSparseLeastSquares}
\end{algorithm}

\section{Related Work}
The fundamental streamlined EB Algorithm
\ref{alg:EMforEBStream} makes use of linear system solutions and sub-block
matrix inverses for two-level sparse matrix problems (\cite{nolan2019solutions}).
Our result for streamlined posterior computation in Section \ref{sec:streamPEB}
is analogous and mathematically identical to Result 2 in
\cite{menictas2019streamlinedCrossed} who instead focus on streamlined
mean field variational Bayes approximate inference for linear mixed models
with crossed random effects. In the present article, we make use
of a similar result for EB posterior inference for use in the
mHealth setting. Our EB algorithm allows streamlined estimation of
the variance components within an online TS contextual bandit algorithm.

Other approaches include
using mixed model software packages for high-performance statistical
computation. For example: \emph{(i)} \texttt{BLME} provides a posteriori estimation
for linear and generalized linear
mixed effects models in a Bayesian setting \cite{dorie2015blme}; and \emph{(ii)}
\texttt{Stan} \cite{carpenter2017stan} provides full Bayesian statistical
inference with MCMC sampling. Even though \texttt{BLME} offers streamlined algorithms
for obtaining the predictions
of fixed and random effects in linear mixed models, the sub-blocks of the covariance matrices
of the posterior required for construction of the EM method in the streamlined empirical
Bayes algorithm are not provided by such software. On the other hand, \texttt{Stan} does
offer support for computation of these sub-blocks, but is well known to suffer
computationally in large data settings.

As we point to in Section \ref{Sec:PerfAssComp}, the Gaussian
linear mixed effects model used in the Thompson-Sampling
algorithm is equivalent to a Gaussian Process regression model
with a structured kernel matrix induced by the use of
random effects. Gaussian process models have been used for
multi-armed bandits
\cite{chowdhury2017kernelized, hoffman2011portfolio, ambikasaran2015fast, srinivas2009gaussian, desautels2014parallelizing, wang2016optimization, djolonga2013high, bogunovic2016time}, and for contextual bandits \cite{li2010contextual, krause2011contextual}.
To address the challenges
posed by mHealth, \cite{tomkins2020intelligentpooling} illustrate the benefits of using
mixed effects Gaussian Process models in the context of reinforcement
learning. Computational challenges in the Gaussian
Process regression setting is a known and common problem
which has led to contributions from the computer
science and machine learning communities. For instance, to
address challenges posed for Gaussian Process regression
suffering from cubic complexity to data size, a variety of
scalable GPs have been presented, including the approach we compare to earlier:
\texttt{GPyTorch} \cite{gardner2018gpytorch}.
A review on state-of-the-art
scalable GPs involving two main categories: global approximations
which distillate the entire data and local approximations
which divide the data for subspace learning can be found in
\cite{liu2018gaussian}. The sparsity imposed by the use of random
effects, however, afford us accurate inference in the
cases considered in this article, and thus do not suffer from
the potential loss of accuracy that could result from the
approximate methods, such as those discussed in \cite{liu2018gaussian}.

\end{document}